\documentclass[letterpaper, 10 pt, conference]{ieeeconf}  

\IEEEoverridecommandlockouts                              

\usepackage{hyperref}

\usepackage{graphics} 
\usepackage{epsfig} 
\usepackage{amsmath} 
\usepackage{color}
\usepackage{hhline}
\usepackage{gensymb}
\newcommand{\comment}[1]{}

\newcommand\blfootnote[1]{%
  \begingroup
  \renewcommand\thefootnote{}\footnote{#1}%
  \addtocounter{footnote}{-1}%
  \endgroup
}
\usepackage{etoolbox}
\makeatletter
\patchcmd{\@makecaption}
  {\scshape}
  {}
  {}
  {}
\makeatletter
\patchcmd{\@makecaption}
  {\\}
  {.\ }
  {}
  {}
\makeatother
\title{\LARGE \bf
End-to-end Trainable Deep Neural Network for Robotic Grasp Detection and Semantic Segmentation from RGB
}

\author{Stefan Ainetter$^{1}$ and Friedrich Fraundorfer$^{1}$
\thanks{$^{1}$All authors are with the Institute of Computer Graphics and Vision,
        Graz University of Technology, 8010 Graz, Austria
        {\tt\small \{stefan.ainetter,fraundorfer\}@icg.tugraz.at}}%
}
\begin{document}

\maketitle
\thispagestyle{empty}
\pagestyle{empty}

\begin{abstract}
In this work, we introduce a novel, end-to-end trainable CNN-based architecture to deliver high quality results for grasp detection suitable for a parallel-plate gripper, and semantic segmentation. Utilizing this, we propose a novel refinement module that takes advantage of previously calculated grasp detection and semantic segmentation and further increases grasp detection accuracy. Our proposed network delivers state-of-the-art accuracy on two popular grasp dataset, namely Cornell and Jacquard. As additional contribution, we provide a novel dataset extension for the OCID dataset, making it possible to evaluate grasp detection in highly challenging scenes. Using this dataset, we show that semantic segmentation can additionally be used to assign grasp candidates to object classes, which can be used to pick specific objects in the scene. 
Source code and dataset extension are available at \url{https://github.com/stefan-ainetter/grasp_det_seg_cnn}.
\blfootnote{The research leading to these results has received funding from the Austrian Ministry for Transport, Innovation and Technology (BMVIT) within the ICT of the Future Programme (4th call) of the Austrian Research Promotion Agency (FFG) under grant agreement n. 864807.} 
\end{abstract}
\section{INTRODUCTION}
\begin{figure} [t]
\centering
        \frame{\includegraphics[width=0.45\linewidth]{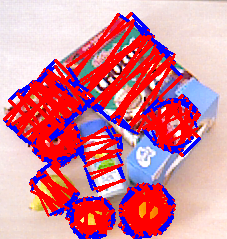}}
 		\frame{\includegraphics[width=0.45\linewidth]{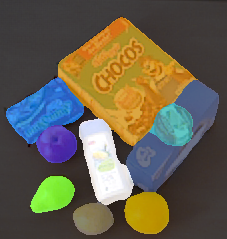}}
      	\frame{\includegraphics[width=0.45\linewidth]{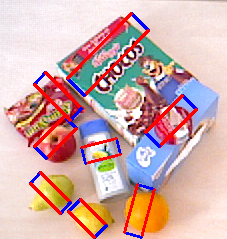}}
 		 \frame{\includegraphics[width=0.45\linewidth]{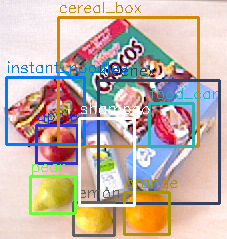}}
\caption[Teaser]{Grasp detection and segmentation results of our proposed model for an image containing multiple graspable objects. \textbf{Top left} shows all predicted grasp candidates in the scene (blue lines denote parallel plates of the gripper, red lines denote opening width), \textbf{top right} shows the predicted semantic segmentation. \textbf{Bottom left} shows that combining both leads to valid grasp candidates for each object in the scene, which makes it possible to decide which object to pick. \textbf{Bottom right} shows the results for object detection represented by axis-aligned bounding boxes. Compared to the semantic segmentation, the bounding box representation is neither suitable to determine which grasp candidate belongs to which object, nor to determine the correct relationship between objects, as boxes highly overlap each other. 
}
\vspace{-12pt}
\label{fig:Teaser}
\end{figure}

Automated grasping is a very active field of research in robotics. The process of having a robot manipulator to successfully grasp objects in a cluttered environment is still a challenging problem. Recent research on robotic grasping led to highly accurate grasp detection in scenes with multiple objects.
Yang et al. \cite{yang2019task} combine grasp detection with a CRF-based semantic model for task-oriented grasping and object detection. The works of \cite{zhang2018multi,park2020multi} combine grasp detection and object detection to predict grasp candidates that are assigned to specific objects in the scene. They showed that object detection provides accurate information about which objects are in the scene, however little information about the object's shape is provided. This can be especially critical in highly stacked scenes, where multiple objects overlap each other.
\\ In this work, we address grasp detection for parallel-plate grippers as well as semantic segmentation, by proposing an end-to-end trainable multi-task deep neural network. We show that one single network is able to provide high quality results for grasp detection and semantic segmentation in challenging scenes. Furthermore, we propose a novel refinement module which combines information about previously calculated grasp candidates and semantic segmentation to further increase the grasp detection accuracy. 
At last, we propose an extension to the OCID dataset \cite{suchi2019easylabel}, by adding hand-annotated grasp candidates and class labels to each object. Evaluation of our proposed model on this OCID dataset extension indicates high accuracy for grasp detection and segmentation in complex scenes with multiple objects. We also show that semantic segmentation can be used to assign grasp candidates to objects, which makes it possible to pick specific objects in the scene. Figure \ref{fig:Teaser} shows results for our novel OCID dataset extension.\\
\vspace{-12pt}
The main contributions of our paper are the following:
\begin{enumerate}
    \item An end-to-end trainable deep neural network architecture for joint grasp detection and dense, pixel-wise semantic segmentation, which yields state-of-the-art performance for grasp detection.
    \item A novel grasp refinement module, combining results from grasp detection and segmentation to further improve the overall grasp accuracy.
    \item An extension of the OCID dataset \cite{suchi2019easylabel} for robotic grasping, by adding ground truth grasp candidates as oriented bounding boxes and class information to each object.
\end{enumerate}
\section{RELATED WORK}
Our work is related to prior research on object detection and semantic segmentation in robotic vision. More specific, it is related to robotic grasp detection and scene understanding for robotic manipulations.
\subsection{Robotic Grasp Detection}
The conventional method for grasp detection uses information about object geometry, physics models and force analytics \cite{bicchi2000robotic}. With the rise of deep learning, data-driven methods \cite{bohg2013data} became more common.
Methods like \cite{redmon2015real,lenz2015deep} were the first to use deep neural networks and supervised learning to predict multiple grasp candidates for a single object. Methods like \cite{chu2018real,zhou2018fully,zhang2018roi,karaoguz2019object,song2020novel} perform grasp detection based on two-stage object detectors. This type of detectors consists of a region proposal network and a detector. After extracting features using proposals from the first stage, objects are detected in the second stage. Approaches like \cite{park2020multi,park2020realtime} are single-stage detectors which divide the input image using a grid and perform detection on each cell. This usually reduces the computation time, with decreased prediction accuracy.\\
Other methods \cite{gkanatsios2020orientation,morrison2018closing} perform grasp detection on depth images as input, where accuracy is highly dependent on the quality of the input data. \\
Kumra et al. \cite{kumra2019antipodal} use a generative architecture to predict pixel-wise grasps in the form of three images, namely quality, angle  and width. Asif et al. \cite{asif2018ensemblenet,asif2018graspnet} use an ensemble of neural networks for grasp detection and segmentation, whereas using multiple networks is usually more expensive in terms of computational cost and memory consumption. DSGD \cite{asif2019densely} uses image information on different levels of the image hierarchy (global-, region-, pixel-level) for grasp detection. 
The works of \cite{yang2019task,zhang2018multi} perform the tasks of object detection with reasoning and grasp detection for complex scenes with multiple objects. 
Other approaches \cite{pinto2016supersizing,levine2018learning} use Reinforcement Learning (RL) on a real or simulated robot to perform thousands of grasp attempts and use the feedback to improve grasp detection. RL has the advantage that no labeled data is necessary for training, but with the downside of being time and hardware consuming.

\subsection{Semantic Segmentation}
Several works use encoder/decoder based CNN architectures \cite{ronneberger2015u,asif2018graspnet} or dilated convolutions \cite{chen2017deeplab} for semantic segmentation. He et al. \cite{he2017mask} performs the task of instance-specific semantic segmentation. In robotic vision, several methods \cite{danielczuk2019segmenting,xie2020best} predict semantic segmentation for unknown objects.
Araki et al. \cite{araki2020mt} proposed a network for semantic segmentation and grasp detection for a suction cup using multi-task learning with a single deep neural network. 
\section{PROBLEM STATEMENT}
\textbf{Grasp detection.}
As proposed by Lenz et al. \cite{lenz2015deep}, we use the five-dimensional rectangular representation for robotic grasps. This representation consists of the location and orientation of a parallel-plate gripper before it closes on the object.
A grasp candidate $\mathbf{g}$ is defined as
\begin{equation}
\mathbf{g}= (x,y,w,h,\theta),
\label{eq:grasp_rep_init}
\end{equation}
whereas $x$ and $y$ describe the center of the grasp candidate, $w$ and $h$ describe the width and height, and $\theta$ describes the orientation of the rotated box representation.
\vspace{2pt}
\\ \textbf{Semantic Segmentation.} Semantic segmentation is the task of assigning a class label to each pixel in the image. Note that for grasp detection of unseen objects, we define the set of semantic classes as $\{graspable, non\text{-}graspable\}$. If additional class information about the objects is available in the dataset, the set of semantic classes can be adjusted accordingly. 
\section{PROPOSED METHOD}
\begin{figure}[htp!] 
      \centering
 		\includegraphics[trim={1cm 0 1cm 0},width=\linewidth]{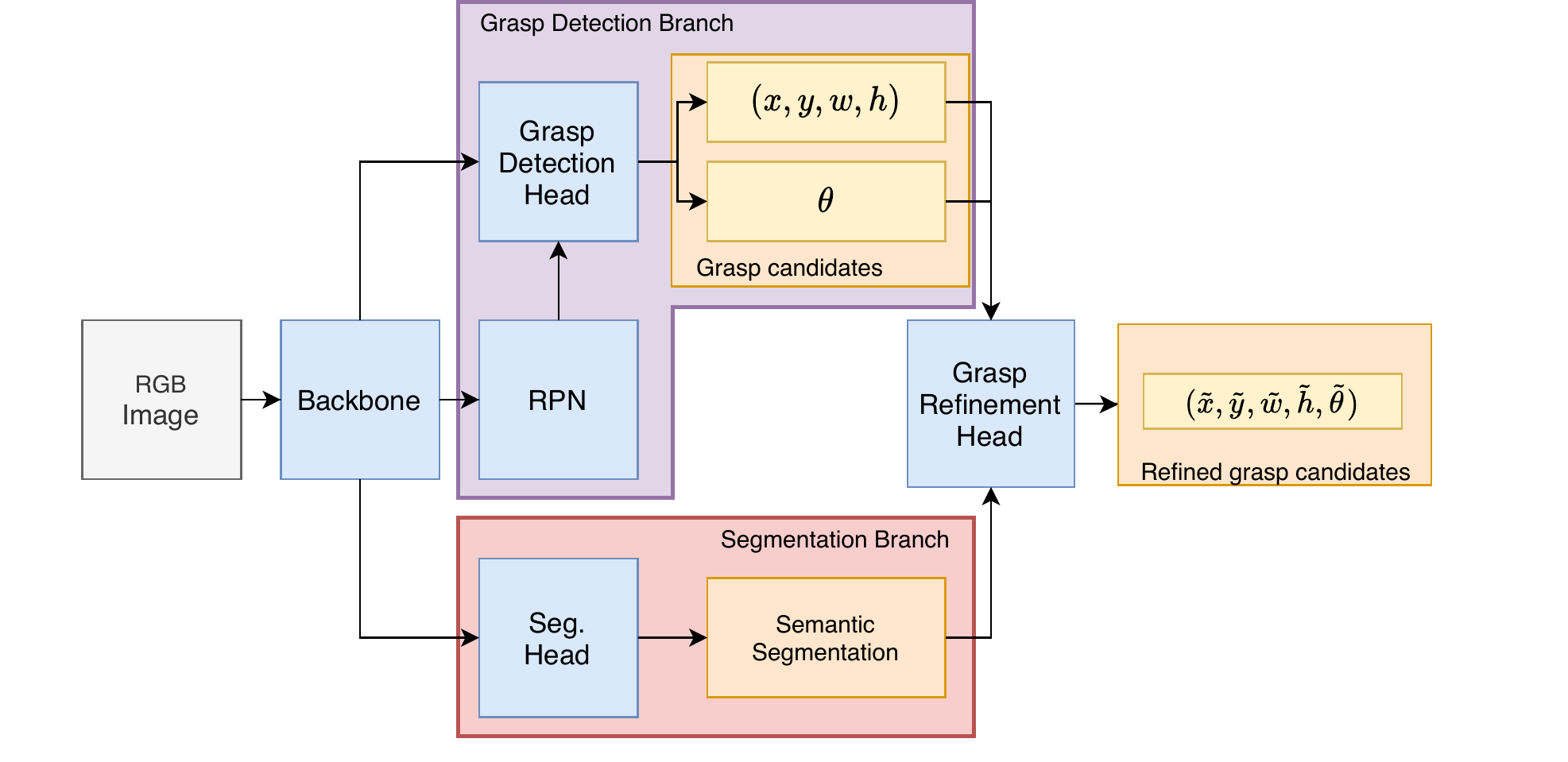}
	\caption[Network architecture]{Architecture of our proposed model. Both branches for grasp detection and segmentation share the backbone network as feature extractor. Both outputs (grasp candidates and semantic segmentation) are used as input for the grasp refinement head, which predicts refined grasp candidates with increased accuracy.}
	\label{fig:Det_Seg_Net}
\end{figure}

The proposed architecture is based on \cite{porzi2019seamless}, a framework originally used for panoptic segmentation, which we adapted for our purpose. This section provides an overview about the specific network parts, which can be seen in Figure \ref{fig:Det_Seg_Net}.
\vspace{-6pt}
\subsection{Shared Backbone}
As feature extractor, we use a slightly modified ResNet-101 \cite{he2016identity} with a Feature Pyramid Network (FPN) \cite{lin2017feature} on top. The modules \{conv2, conv3, con4, conv5\} of the ResNet-101 architecture are linked to the FPN. All Batch Normalization + ReLU layers in the original ResNet-101 structure are replaced by synchronized Inplace Activated Batch Normalization (InPlaceABNSync) \cite{rota2018place}, using LeakyReLU as activation function. 
\subsection{Grasp Detection Branch}
The grasp detection branch is based on the state-of-the-art Faster R-CNN object detector \cite{ren2015faster}, consisting of a Region Proposal Network (RPN) and a detection stage. This subsection provides details about how we modified this object detector for grasp detection.
\subsubsection{Region Proposal Network}
The RPN outputs rectangular region proposals using the features of the backbone network as input. These region proposals are defined as $\mathbf{\hat{r}}=(\hat{x},\hat{y},\hat{w},\hat{h})$ where $(\hat{x},\hat{y})$ denote the center of the proposal in pixel coordinates and $(\hat{w},\hat{h})$ denote the width and height dimension, respectively. 
Note that the region proposals are axis-aligned, without information about a possible orientation.
\subsubsection{Grasp Detection Head}
The grasp detection head predicts grasp candidates, whereas one candidate $\mathbf{g}$ is defined as described in Equation \ref{eq:grasp_rep_init}. Each previously calculated region proposal $\mathbf{\hat{r}}$ is used as input for the grasp detection head. Then, ROIAlign \cite{he2017mask} is applied to extract feature maps with $14 \times 14$ spatial resolution that directly correspond to the region proposals. Afterwards, average pooling with kernel size 2 is applied to each feature map, before they are fed into two fully connected (fc) layers with $1024$ neurons each. After each fc layer, an InPlaceABNSync normalization layer and Leaky ReLU activation with $0.01$ is applied. The results are forwarded into two sub-networks:
	\vspace{2pt}
\\ \textbf{Grasp orientation prediction.} The first sub-network consists of a fc layer with $1024$ neurons followed by $N_{classes} + 1$ output units, whereas $N_{classes}$ defines the number of orientation classes. Similar to \cite{chu2018real}, we discretized the grasp orientation $\theta$ into $N_{classes} = 18$ intervals with equal length, where each interval is represented by its mean value. The full set of possible orientation classes is defined as $\mathcal{C} = \{(1,...,N_{classes})\}$ and the additional class $\emptyset$ to asses the possible invalidity of a proposal. 
The output units provide logits for a softmax layer that represents the probability distribution over all possible orientation classes.
The probability associated to orientation class $c \in \mathcal{C}$ is used as score function $s^{c}$.
	\vspace{2pt}
\\ \textbf{Bounding box prediction.}
The second sub-network consists of a fc layer with $1024$ neurons followed by $4 N_{classes}$ output units. The output units encode class-specific correction factors $(t^{c}_{x},t^{c}_{y},t^{c}_{w},t^{c}_{h})$ for each input proposal $\mathbf{\hat{r}}$. \\ These correction factors, and the orientation information given by the score function $s^{c}$, can then directly be used to calculate a grasp candidate $\mathbf{g}$. 
\subsection{Segmentation Branch}
The segmentation head takes the output of the first four scales of the FPN as input. Each scale of FPN features are fed to an independent Mini-DeepLab module \cite{porzi2019seamless}, which makes it possible to capture global structures of the input image with relatively few memory consumption. The output of each Mini-DeepLab module is then up-sampled to $\frac{1}{4}$ of the input image size. Afterwards, all feature maps are concatenated and fed to a $1 \times 1$ convolution with $S_{classes}$ output channels, representing the probability distribution over all semantic classes.

\subsection{Grasp Refinement Head}
\begin{figure}[htp!] 
      \centering
 		\includegraphics[trim={0 0 0cm 0.4cm},width=\linewidth]{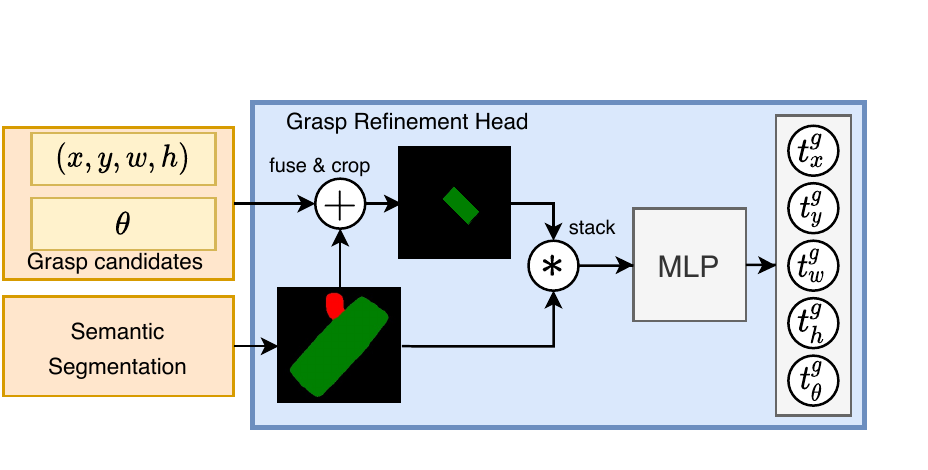}
	\caption[Network architecture]{Structure of the grasp refinement head. The Multilayer Perceptron (MLP) takes as input two stacked semantic probability maps with dimension $(H \times W)$, whereas $(H,W)$ defines the height and width of the probability map. The output units of the MLP represent refined correction factors $(t_{x}^{g},t_{y}^{g},t_{w}^{g},t_{h}^{g},t_{\theta}^{g})$ which can directly be used to calculate the refined grasp candidate. Note that this operation is performed simultaneously for $N$ grasp candidates, leading to an $(N \times 2 \times H \times W)$ dimensional input for the MLP.}
	\label{fig:Refine_module}
\end{figure}

The grasp refinement head (see Figure \ref{fig:Refine_module}) takes as input the previously predicted grasp candidates from the grasp detection branch and a semantic probability map from the segmentation branch. First, we fuse both information by cropping the area of the grasp candidate from the probability map. The cropped probability map and the original probability map are then stacked together and used as input for a Multilayer Perceptron (MLP). Note that this approach enables us to combine geometric information of grasp candidates with information about object shape. 
The MLP is a two-layer fc network with five output neurons. After each fc layer, an InPlaceABNSync normalization layer and Leaky ReLU activation with $0.01$ is applied. The output represents refined correction factors $(t^{g}_{x},t^{g}_{y},t^{g}_{w},t^{g}_{h},t^{g}_{\theta} )$ for each input grasp candidate $\mathbf{g}$, which are then used to calculate the refined grasp candidate $\mathbf{\tilde{g}}$.
\vspace{-6pt}
\section{Training Losses}
For simultaneously learning the tasks of grasp detection, instance segmentation and grasp candidate refinement, we define the composite loss function $\mathcal{L}$ as
\begin{equation}
\mathcal{L}= \lambda_{grasp}\mathcal{L}_{grasp} + \lambda_{sem}\mathcal{L}_{sem} + \lambda_{refine}\mathcal{L}_{refine},
\label{eq:LOSS_full}
\end{equation}
with the grasp detection loss $\mathcal{L}_{grasp}$, the semantic segmentation loss $\mathcal{L}_{sem}$ and the grasp refinement loss $\mathcal{L}_{refine}$. All parts are weighted with a specific hyper-parameter $\lambda$. \\ 
\subsection{Grasp Detection Loss}
The grasp detection loss $\mathcal{L}_{grasp}$ is defined as
\begin{equation}
\mathcal{L}_{grasp}= \mathcal{L}_{RPN} + \mathcal{L}_{box} + \mathcal{L}_{rot},
\label{eq:LOSS_grasp}
\end{equation}
where $\mathcal{L}_{RPN}$ defines the loss for training the RPN, $\mathcal{L}_{box}$ defines the regression loss for the box coordinates and $\mathcal{L}_{rot}$ defines the classification loss for the grasp orientation.
We refer to \cite{he2017mask} for additional information about $\mathcal{L}_{RPN}$. \\ 
The grasp orientation loss $\mathcal{L}_{rot}$ is defined as
\begin{equation}
\mathcal{L}_{rot} = - \frac{1}{\mathcal{|R|}} \sum_{\hat{r} \in \mathcal{R_{+}}} \log s_{\hat{r}}^{c_{r}}   - \frac{1}{\mathcal{|R|}} \sum_{\hat{r} \in \mathcal{R_{-}}} \log s_{\hat{r}}^{\emptyset}.
\label{eq:rot_loss}
\end{equation}
Note that $\mathcal{R} = \mathcal{R_{+}} \cup \mathcal{R_{-}}$ is the set of valid and invalid region proposals generated using the RPN. The score function $s_{\hat{r}}^{c_{r}}$ defines the probability that the region proposal $\mathbf{\hat{r}}$ belongs to the ground truth orientation class $c_{r}$, and $s_{\hat{r}}^{\emptyset}$ defines the probability that the region proposal is invalid. \\
For bounding box regression we use the loss $\mathcal{L}_{box}$ defined as
\begin{equation}
\mathcal{L}_{box} = \sum_{i \in \{x,y,h,w\}} smooth_{L_{1}} (t_{i}^{c} - t_{i}^{*}), 
\label{eq:box_loss}
\end{equation}
with the $smooth_{L_{1}}$ norm \cite{ren2015faster} defined as 
\begin{equation}
 smooth_{L_{1}}(z) = 
\begin{cases}
    0.5z^{2},& \text{if } |z| < 1\\
    |z| - 0.5,              & \text{otherwise}.
\end{cases}
\label{eq:smoothL1}
\end{equation}
Note that the parameters $t_{i}^{c}$ are the output of the bounding box prediction.
The correction factors $t_{i}^{*}$ \cite{girshick2014rich} are defined  as:
\begin{equation}
\label{eq:box_reg}
\begin{aligned}
t_{x}^{*} = (x^{*} - \hat{x}) / \hat{w}, \quad \quad t_{y}^{*} = (y^{*} - \hat{y}) /\hat{h} \\
t_{w}^{*} = \log (w^{*}/\hat{w}), \quad \quad t_{h}^{*} = \log (h^{*}/\hat{h}),
\end{aligned}
\end{equation}
whereas
variables $(\hat{x},\hat{y},\hat{w},\hat{h})$ and $(x^{*},y^{*},w^{*},h^{*})$ are parameters from the region proposal $\mathbf{\hat{r}}$ and ground truth grasp candidate respectively.
\subsection{Segmentation Loss}
We denote $\mathcal{Y} = \{1,..., S_{classes}\}$ as the set of semantic segmentation classes. The per-image segmentation loss is a weighted per-pixel loss \cite{porzi2019seamless} defined as
\begin{equation}
\mathcal{L}_{sem} = - \sum_{j,k} w_{j,k} \log P_{j,k}(Y_{j,k}),
\label{eq:sem_loss}
\end{equation}
where $(j,k)$ correspond to the pixel position in the image. Let $Y_{j,k} \in \mathcal{Y}$ be the semantic segmentation ground truth and $P_{j,k}(s)$ the predicted probability for the same pixel to be assigned to a semantic class $s \in \mathcal{Y}$, respectively. The weights $w_{j,k}$ are computed to implement a pixel-wise hard negative mining, which selects the $25\%$ of the lowest predicted probabilities $P_{j,k}$ for all $(j,k)$ using $w_{j,k} = \frac{4}{WH}$, and $w_{j,k} = 0$ otherwise. The spatial resolution of the input image is given as $(W \times H)$.
\subsection{Grasp Refinement Loss}
The grasp refinement loss $\mathcal{L}_{refine}$ is defined as
\begin{equation}
\mathcal{L}_{refine} = \sum_{i \in \{x,y,h,w,\theta\}} smooth_{L_{1}} (t^{g}_{i} - \tilde{t}^{*}_{i}). 
\label{eq:refine_loss}
\end{equation}
Compared to $\mathcal{L}_{box}$ in Equation \ref{eq:box_loss}, the smooth L1 loss is additionally calculated for the orientation parameter $\theta$. Note that the parameters $t^{g}_{i}$ are the output of the grasp refinement head. The correction factors  $\tilde{t}^{*}_{i}$ were adapted based on \cite{Ding_2019_CVPR}, to take the orientation of the previously predicted grasp candidate $\mathbf{g}$ into account:
\begin{equation}
\label{eq:box_reg_refine}
\begin{aligned}
\tilde{t}_{x}^{*} = \frac{1}{w}((x^{*} - x) \cos \theta + (y^{*} - y) \sin \theta) \\ 
\tilde{t}_{y}^{*} = \frac{1}{h}((y^{*} - y) \cos \theta - (x^{*} - x) \sin \theta) \\
\tilde{t}_{w}^{*} = \log (w^{*}/w), \quad \quad \tilde{t}_{h}^{*} = \log (h^{*}/h) \\
\tilde{t}_{\theta}^{*} = \frac{1}{\pi}((\theta^{*} - \theta) \; mod \; \pi ).
\end{aligned}
\end{equation}
Note that variables $(x,y,w,h,\theta)$ and $(x^{*},y^{*},w^{*},h^{*},\theta^{*})$ are parameters from the grasp candidate $\mathbf{g}$ and the ground truth grasp candidate respectively. The $mod$ operation is used to adjust the orientation offset $\tilde{t}_{\theta}^{*}$ to $[0,\pi)$, which corresponds to the interval of possible values for the grasp orientation.
\section{Experiments and Results}
We assess the benefits of our proposed method on well-known grasping datasets, namely Cornell \cite{lenz2015deep} and Jacquard \cite{depierre2018jacquard}. Additionally, we evaluate our approach on our novel OCID \cite{suchi2019easylabel} extension for robotic grasping. We compare our proposed architecture to state-of-the-art methods which also only use RGB data as input modality, to ensure fair comparison. Due to the fact that our grasp detection and segmentation branches are independent, we evaluate several model configurations throughout our experiments, with the following terminology: Detection (ours) refers to the model using only the grasp detection branch, Det\_Seg (ours) consists of both grasp detection and segmentation branch, and Det\_Seg\_Refine (ours) consists of the full architecture (see Figure \ref{fig:Det_Seg_Net}). We initialize the backbone network with pre-trained ImageNet \cite{russakovsky2015imagenet} weights and freeze the parameters of the first two network modules \{conv1, conv2\} during all training runs. 
All training and evaluation runs were performed using a single Nvidia GeForce RTX 2080 Ti graphics card. Unless otherwise stated, results from other methods are taken from the corresponding paper.

\subsection{Evaluation Metric}
\label{seq:eval_metric}
For grasp detection, we use the popular Jaccard index as accuracy measurement. A grasp candidate is reported correct if:
\begin{enumerate}
    \item the angle difference between predicted grasp candidate $g_{p}$ and ground truth grasp candidate $g_{gt}$ is within $30^{\circ}$ and
    \item the Intersection over Union (IoU) of them is greater than $0.25$ which means 
    \begin{equation}
        IoU = \frac{\lvert g_{p} \cap g_{gt} \rvert}{\lvert g_{p} \cup g_{gt} \rvert} > 0.25
    \end{equation}
\end{enumerate}
For semantic segmentation, we report the IoU between predicted and ground truth segmentation.

\subsection{Cornell dataset}
The Cornell grasp dataset \cite{lenz2015deep} contains $885$ RGB-D images of $240$ graspable objects, whereas each image has a dimension of  $640 \times 480px$. Correct grasp candidates are hand-annotated using the rectangular grasp representation. Like previous work, we used 5-fold cross-validation and report the mean grasp detection accuracy. We performed an image-wise split into training and test set, whereas image-wise means that images are split randomly without considering which object is in the image.
\comment{Note that we only use the RGB data, and therefore evaluate against methods with the same data input which guarantees fair comparison.}
Due to the fact that this dataset does not provide ground truth segmentation data, we are only able to evaluate the model Detection (ours) in this experiment.
\\ \textbf{Data Preprocessing and Data Augmentation.}
Because of the relatively small size of this dataset, we used data augmentation to enlarge the dataset during training. Each image was center cropped to $351 \times 351px$. Next, we applied random rotation between $0^{\circ} \text{ and } 360^{\circ}$, and randomly translated the image in $x$ and $y$ direction independently up to $50px$.
\\ \textbf{Training schedule.}
The network was trained end-to-end, using a learning rate of 0.04 with weight decay of 0.0001, a momentum factor of 0.9 with enabled nesterov momentum and SGD as optimizer. We used a batch size of 12 during training.
\\ \textbf{Quantitative Results.}
Table \ref{tab:detection_cornell_rgb} shows the results on the Cornell grasp dataset. Note that our implementation achieves state-of-the-art results for accuracy compared to other methods using RGB images as input. Additionally, our model achieves a high FPS rate during inference, making it well suited for real-time applications.
\begin{table}[htp!]
\centering
\caption{Comparison of grasp accuracy and computational speed for the Cornell dataset. All results are produced with image-wise data split and RGB as input modality using 5-fold cross-validation.}
\label{tab:detection_cornell_rgb}
\begin{tabular}{|c|c|c|c|}
\hline
Method                       & Input &\begin{tabular}[c]{@{}c@{}}Grasp \\ Accuracy (\%) \end{tabular} & \begin{tabular}[c]{@{}c@{}}Speed\\ (FPS)\end{tabular} \\ \hline
Zhou  \cite{zhou2018fully}, Resnet-50              & RGB   & 97.7     & 9.9      \\ \hline
Zhou  \cite{zhou2018fully}, Resnet-101             & RGB   & 97.7    & 8.5       \\ \hline
Zhang \cite{zhang2018roi}, Resnet-101            & RGB   & 93.6    & 25.2      \\ \hline
Park \cite{park2020multi}, Darknet-19             & RGB   & 97.7    & \textbf{140}    \\ \hline
Karaoguz \cite{karaoguz2019object}, VGG-16             & RGB   & 88.7    & 2       \\ \hline
Chu \cite{chu2018real}, Resnet-50               & RGB   & 94.4      & 8.3   \\ \hline
Kumra \cite{kumra2019antipodal}, GR-ConvNet            & RGB   & 96.6     & 52       \\  \hhline{|=|=|=|=|}
Detection (ours) & RGB   & \textbf{98.2}         & 63  \\ \hline
\end{tabular}
\end{table}
\vspace{-6pt}
\subsection{Jacquard Dataset}
The Jacquard dataset \cite{depierre2018jacquard} contains 54k synthetic RGB-D images using 11k different objects. As ground truth, the dataset contains automatically generated grasp candidates, as well as ground truth semantic segmentation, which enables us to evaluate our multi-task, grasp detection and segmentation approach. We made an image-wise split of the dataset, using $95\%$ of the data as training set, and the remaining images as test set.
\\ \textbf{Data Preprocessing.}
Due to the relatively large image size of $1024 \times 1024px$ , we decided to downsample each image by the factor of 2. No data augmentation was needed due to the high number of samples in the dataset.
\\ \textbf{Training schedule.}
For this experiments, the networks Detection (ours) and Det\_Seg (ours) were trained end-to-end using a learning rate of 0.023 with weight decay of 0.0001, a momentum factor of 0.9 with enabled nesterov momentum and SGD as optimizer. We used a batch size of 8 during training and weighting factors of $\lambda_{grasp}= 1.0$, $\lambda_{sem}=0.8$ and $\lambda_{refine}=0.8$ for loss balancing.
For training Det\_Seg\_Refine (ours), we started from the pre-trained version Det\_Seg (ours), freezing all backbone parameters while re-training all other parts. We changed the learning rate to 0.008 and the batch size to 4. Although Det\_Seg\_Refine (ours) could also be trained from scratch, we believe that starting from a well-known baseline makes it easier to demonstrate the effect of the proposed extension and lead to results that are easier to interpret.
\\ \textbf{Quantitative Results.}
Our experiments on the Jacquard dataset lead to several findings:
\paragraph{All of our model configurations deliver state-of-the-art performances} Table \ref{tab:detection_jacquard_rgb} shows the results on the Jacquard grasp dataset according to grasp accuracy, segmentation accuracy and computational speed.  All our models achieve higher scores for grasp detection accuracy compared to other state-of-the-art methods using RGB images as input. Furthermore, the inference speed indicates that all our models are fast enough for real-time application.
\paragraph{Multi-task learning is not automatically beneficial} Focusing on the results of Det\_Seg (ours), we see that multi-task learning for grasp detection and segmentation has in our case no beneficial effect on grasp accuracy. However, given the fact that these two tasks are quite different (grasp detection focuses on very small parts of the object whereas segmentation takes the whole scene into account) it is already remarkable that both models perform nearly equally well.
\paragraph{The grasp refinement head significantly increases orientation accuracy} Table \ref{tab:detection_jacquard_rgb} shows that using Det\_Seg\_Refine (ours) increases the grasp accuracy by a small margin compared to Det\_Seg (ours). 
Due to the fact that the Jaccard index criteria are quite weak (see Subsection \ref{seq:eval_metric}), Det\_Seg (ours) is accurate enough to pass them in most cases. However, if we apply more strict criteria, we can see a significant benefit using our grasp refinement module. Table \ref{tab:jacquad_angle_diff} shows the results for decreased orientation tolerance. Using our proposed refinement head, we were able to outperform our baseline model by more than $16\%$, which highlights a significant increase of orientation accuracy (see Table \ref{tab:jacquad_angle_diff}, column $5\degree$).
Table \ref{tab:jacquard_IoU_diff} shows the results for different IoU criteria. As there is no significant difference for our two models, this experiment indicates that multiple iterations of bounding box regression do not increase the performance (as stated in \cite{girshick2014rich}). 

\begin{table}[htp!]
\centering
\caption{Comparison of grasp accuracy, segmentation IoU and computational speed for the Jacquard dataset. All results are produced with image-wise data split and RGB as input modality.}
\label{tab:detection_jacquard_rgb}
\begin{tabular}{|c|c|c|c|}
\hline
Method                        & \begin{tabular}[c]{@{}c@{}}Grasp \\ Accuracy (\%) \end{tabular} & \begin{tabular}[c]{@{}c@{}}Seg. \\ IoU (\%) \end{tabular} & \begin{tabular}[c]{@{}c@{}}Speed\\ (FPS)\end{tabular} \\ \hline
Zhang\cite{zhang2018roi}, ROI-GD      & 90.4  & -           & -     \\ \hline
Song \cite{song2020novel}, Resnet-101              &91.5         & -   & -     \\ \hline
Kumra \cite{kumra2019antipodal}, GR-ConvNet             & 91.8       & -       & -     \\ \hline
Depierre \cite{depierre2020optimizing}                      & 85.7    & -   & -     \\ \hhline{|=|=|=|=|}
Detection (ours)  & 92.69                & -                & \textbf{53}     \\ \hline
Det\_Seg (ours) & 92.59 & \textbf{95.12}             & 48     \\ \hline
Det\_Seg\_Refine (ours) &  \textbf{92.95}  & 94.86             & 31     \\ \hline
\end{tabular}
\end{table}
\begin{table}[]
\centering
\caption{Comparison of grasp accuracy (in [\%]) for the Jacquard dataset using different angle thresholds and an IoU threshold of $25\%$. Results for the methods Zhou and Depierre are taken from \cite{depierre2020optimizing}.}
\label{tab:jacquad_angle_diff}
\begin{tabular}{|c|c|c|c|c|c|c|}
\hline
Method                    &  30 \degree  & 25 \degree  & 20 \degree  & 15 \degree  & 10 \degree & 5 \degree \\ \hline
Zhou \cite{zhou2018fully} & 81.95        & 81.76       & 81.27       & 80.23    & 77.79     & -   \\ \hline
Depierre \cite{depierre2020optimizing} & 85.74        & 85.55       & 85.01       & 83.65    & 80.82     & -    \\  \hhline{|=|=|=|=|=|=|=|}
Detection (ours)          & 92.69        & 92.34       & 92.08       & 91.40    & 88.12     & 56.23  \\ \hline
\begin{tabular}[c]{@{}c@{}} Det\_Seg\_Refine \\ (ours) \end{tabular} &   \textbf{92.95}      &  \textbf{92.88}      & \textbf{92.42}  & \textbf{91.52} & \textbf{88.92} &  \textbf{72.79}     \\ \hline

\end{tabular}
\end{table}

\begin{table}[]
\centering
\caption{Grasp accuracy (in [\%]) for the Jacquard dataset using different IoU thresholds and an angle threshold of $30\degree$. Results for the methods Zhou and Depierre are taken from \cite{depierre2020optimizing}.}
\label{tab:jacquard_IoU_diff}
\begin{tabular}{|c|c|c|c|}
\hline
Method                    & IoU 25\% & IoU 30\% & IoU 35\% \\ \hline
Zhou \cite{zhou2018fully} & 81.95    & 78.26    & 74.33   \\ \hline
Depierre \cite{depierre2020optimizing}   & 85.74    & 82.58    & 78.71   \\ \hline
Song \cite{song2020novel} & 91.5     & 89.7     & 87.3    \\  \hhline{|=|=|=|=|}
Detection (ours)          & 92.69    & 91.29    & \textbf{88.99}    \\ \hline
\begin{tabular}[c]{@{}c@{}} Det\_Seg\_Refine \\ (ours) \end{tabular} & \textbf{92.95}  &  \textbf{91.33}  &  88.96   \\ \hline
\end{tabular}
\end{table}
\subsection{OCID Dataset with Grasp-Extension}
\begin{figure}[htp!] 
      \centering
 		\includegraphics[trim={0 0 0cm 0},width=\linewidth]{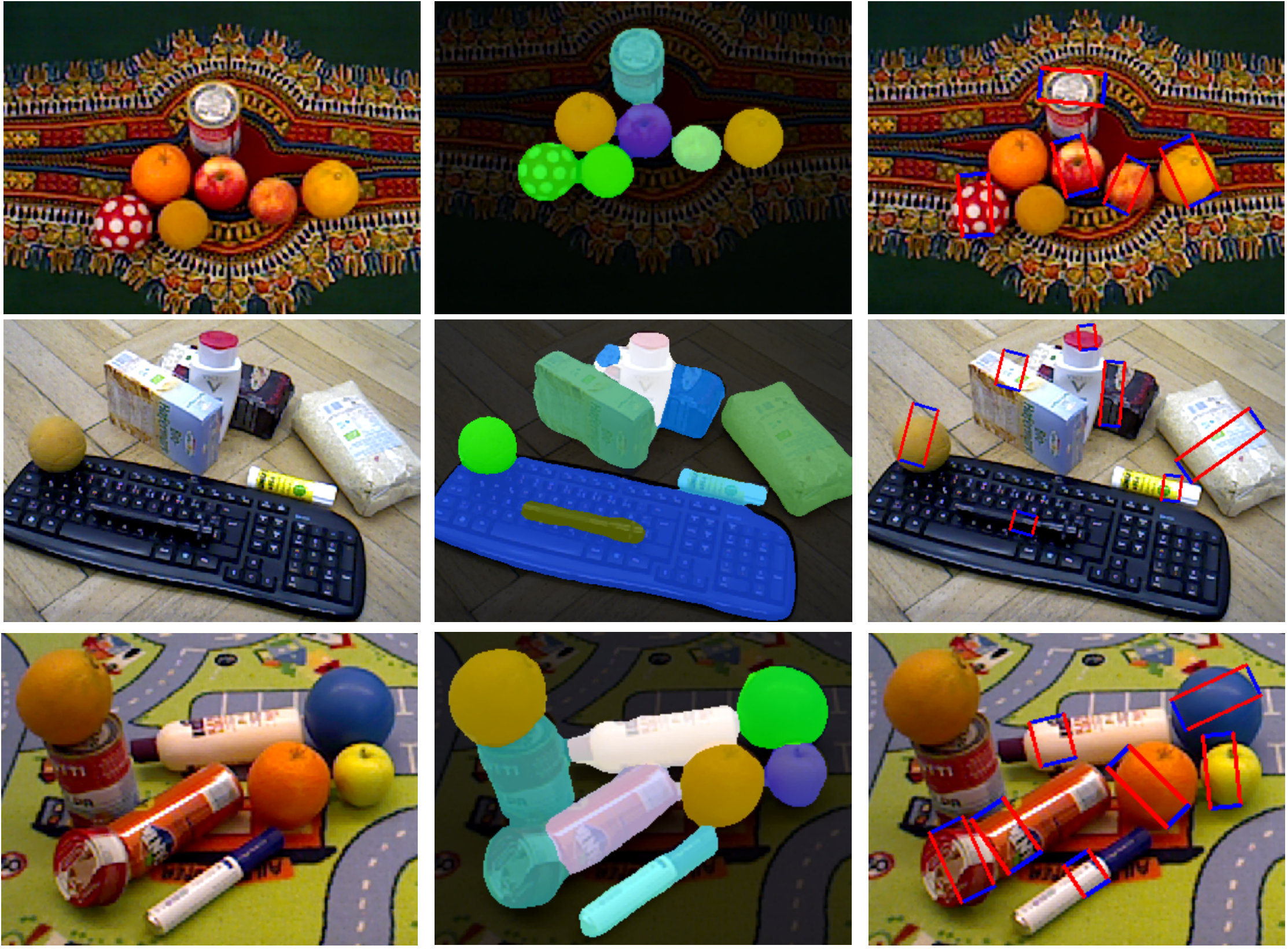}
	\caption[]{Qualitative results for OCID\_grasp. Explanation of images from left to right: 1) raw input image; 2) predicted semantic segmentation, whereas each color represents a specific class; 3) best possible grasp for each class in the scene (blue lines denote parallel plates of the gripper, red lines denote opening width). If multiple instances of the same class are in the scene, e.g. class orange in the first and third row, only the one grasp candidate with the highest confidence will be shown. Each row refers to a separate input example.}
	\label{fig:OCID_full}
\end{figure}

OCID \cite{suchi2019easylabel} contains diverse settings of objects, background, context, sensor to scene distance, viewpoint angle and lighting conditions. The original purpose of this dataset is to evaluate semantic segmentation methods in scenes with increasing amount of complexity. OCID contains RGB-D data with automatically annotated segmentation. For our proposed dataset extension (we denote it as OCID\_grasp), we manually annotated the ARID10 and ARID20 subsets of OCID with valid grasp candidates for each graspable object in the scene. Additionally, we added class information to each object, and assigned each grasp candidate to the corresponding object class. This allows us to evaluate grasp candidates for each object class in the scene.
Overall, OCID\_grasp consists of 1763 selected images with over 11.4k segmented object masks and more than 75k hand-annotated grasp candidates, and each object is classified into one of $31$ classes.
Figure \ref{fig:OCID_full} (left column) shows sample images of the dataset, indicating the high complexity of the scenes.
\\ \textbf{Data Preprocessing and Augmentation.}
We applied random rotation between $0^{\circ} \text{ and } 360^{\circ}$, and randomly translated the image in x and y direction independently up to $50px$ as data augmentation during training.
\\ \textbf{Training schedule.}
The network was trained end-to-end using a learning rate of 0.03 with weight decay of 0.0001, a momentum factor of 0.9 with enabled nesterov momentum and SGD as optimizer. We used a batch size of 8 during training and weighting factors of $\lambda_{grasp}= 1.0$ and $\lambda_{sem}=0.75$ for loss balancing.
\\ \textbf{Evaluation metrics.}
Due to the fact that the scenes in this dataset contain multiple graspable objects, we decided to calculate the Jaccard index for each graspable object class in the scene, as well as the segmentation IoU for each object class.
The Jaccard index at object class level has the additional criteria that the center of the grasp candidate has to be located on the predicted segmentation mask of this specific class. We used 5-fold cross-validation and report the mean grasp detection and segmentation accuracy.
\\ \textbf{Quantitative \& Qualitative Results.}
We used Det\_Seg (ours) as network architecture for this experiment. The network outputs possible grasp candidates for each graspable object in the scene, as well as semantic segmentation for each class. We selected the best grasp candidate for each object class according to the grasp confidence score, with the constraint that the center of the grasp candidate has to be located on the segmentation mask of this object class. Table \ref{tab:OCID_results} shows the results for grasp accuracy at object class level and mean IOU for instance segmentation. Both, the high grasp accuracy and segmentation IoU, indicate that our proposed model delivers highly accurate results for complex scenes. Qualitative results are shown in Figure \ref{fig:OCID_full}. 
\begin{table}[]
\centering
\caption{Results for grasp accuracy and semantic segmentation for the OCID\_grasp dataset. For grasp accuracy, the Jaccard index was calculated on object class level, which corresponds to finding the best possible grasp for each class.}
\label{tab:OCID_results}
\begin{tabular}{|c|c|c|}
\hline
Method          & Grasp Accuracy (\%)  & Seg. IoU (\%) \\ \hline
Deg\_Seg (ours) & 89.02  & 94.05   \\ \hline
\end{tabular}
\end{table}
\section{CONCLUSIONS}
In this work, we have proposed a novel CNN architecture for jointly producing highly accurate grasp detection and semantic segmentation, using a shared backbone network as feature extractor. We showed that our proposed grasp refinement module can be successfully used to increase the accuracy of previously predicted grasp candidates, especially in terms of grasp orientation. Using our proposed OCID\_grasp dataset extension, we showed high accuracy for grasp detection in complex scenes, and demonstrated how semantic segmentation can additionally be used to assign grasp candidates to specific objects. In the future, we plan to extend our approach using multi-modal input data to leverage the full extend of information provided by RGB-D grasp datasets.


\bibliographystyle{IEEEtran} 
\bibliography{./IEEEabrv,./IEEEexample}
\end{document}